\documentclass[letterpaper, 10 pt, conference]{ieeeconf}  
\IEEEoverridecommandlockouts                              
\usepackage{graphics} 
\usepackage{epsfig} 
\usepackage{mathptmx} 
\usepackage{times} 
\usepackage{amsmath} 
\usepackage{amssymb}  
\usepackage{siunitx}
\usepackage{lipsum}
\usepackage{dblfloatfix}  
\usepackage{float}
\DeclareSIUnit{\bodylength}{BL}
\title{\LARGE \bf Snapping Actuators with Asymmetric and Sequenced Motion}

\author{Xin Li, Ye Jin, Mohsen Jafarpour, Hugo de Souza Oliveira, Edoardo Milana
\thanks{All authors are with the Department of Microsystems Engineering (IMTEK) and the Cluster of Excellence livMatS @ FIT – Freiburg Center for Interactive Materials and Bioinspired Technologies, University of Freiburg, Freiburg, Germany}
\thanks{Corresponding author: edoardo.milana@imtek.uni-freiburg.de}
}

\begin{document}

\maketitle
\thispagestyle{empty}
\pagestyle{empty}


\begin{abstract}
Snapping instabilities in soft structures offer a powerful pathway to achieve rapid and energy-efficient actuation. In this study, an eccentric dome-shaped snapping actuator is developed to generate controllable asymmetric motion through geometry-induced instability. Finite element simulations and experiments reveal consistent asymmetric deformation and the corresponding pressure characteristics. By coupling four snapping actuators in a pneumatic network, a compact quadrupedal robot achieves coordinated wavelike locomotion using only a single pressure input. The robot exhibits frequency-dependent performance with a maximum speed of 72.78~mm\,s\(^{-1}\) at 7.5~Hz. These findings demonstrate the potential of asymmetric snapping mechanisms for physically controlled actuation and lay the groundwork for fully untethered and efficient soft robotic systems.
\end{abstract}


\section{Introduction}

In the last decades, soft robotics has been extensively investigated from different perspectives, including materials, structural design, compliant mechanisms, and advanced control algorithms. The compliance and adaptability of soft robotic systems make them particularly well suited for real-world applications, especially tasks that involve close interaction with humans or unstructured environments in general.

In most cases, the control architecture of inflatable soft robots relies on valves, pumps, and micro controllers, which are often bulky and tethered to external power supplies. Such designs limit the robot’s mobility and make it difficult to integrate the control components directly into the body of the robot. To overcome this limitation, the concept of physical control, a principle inspired by biological systems in nature, was explored \cite{milana_santina_gorissen_rothemund_2025}. This approach seeks to embed functional “intelligence” leveraging the physics of soft materials and structures that compose the body of soft robots, enabling them to exhibit complex behaviors without relying on centralized control.

One particularly promising mechanism to enable physical control are snapping structures. Snapping structures possess specially designed geometries that allow them to store elastic energy and release it rapidly when a snapping threshold is crossed. This phenomenon is ubiquitous in both nature and everyday life.

Several approaches have demonstrated the potential of snapping structures in soft robotics \cite{pal_restrepo_goswami_martinez_2021}. For instance, Gorissen et al. developed an inflatable snapping shell for a jumping robot, achieving a jump height of \SI{42.9}{\milli\metre} with a slow inflation rate of \SI{10}{\milli\litre\per\minute}. Similarly, Qiao et al.~\cite{qiaoBiShellValveFast2021a} designed a snapping shell valve that couples a spherical cap with a bistable shell to convert a slow input into a rapid volumetric output. Comoretto et al. \cite{comorettoEmbodyingMechanofluidicMemory2025} integrated mechanofluidic memory into soft robots, using bistable shells and soft tubes to achieve autonomous behavior switching without electronic control. These examples demonstrate how snapping mechanisms can enable fast actuation, energy-efficient motion, and compact integration of control functions within the robot body itself.
\begin{figure}[t]
    \centering
    \includegraphics[width=0.4\textwidth]{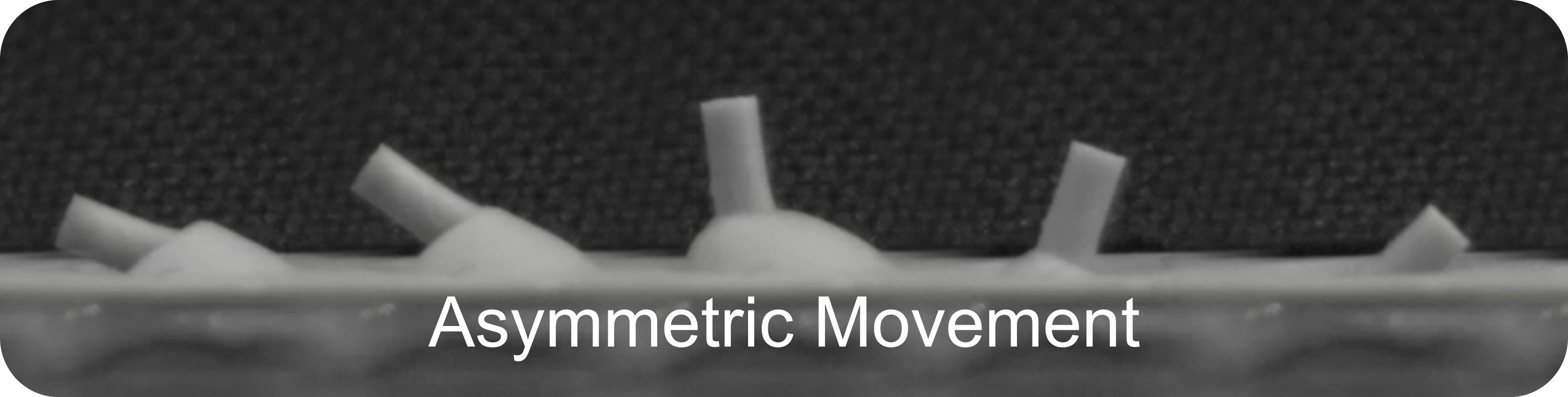}
    \caption{Asymmetric movement of the eccentric snapping actuator.}
    \label{fig:fig0}
\end{figure}

While most approaches to snapping structures emphasize their energy storage and rapid release capabilities, the role of asymmetric motion during snap-through transitions has received comparatively less attention. Recent advances in soft robotics demonstrate that introducing asymmetries—whether in geometry, material distribution, or actuation—can deterministically steer these transitions~\cite{bell_cattani_gorissen_bertoldi_weaver_wood_2020,Li_2024,Luo_2024,Osorio_2025}. Such strategies facilitate directional, programmable, and adaptive mechanical responses, supporting robust and multifunctional behaviors in soft robotic systems.

Asymmetric design is central to controlling snap-through in soft membranes and shells. Modifications such as localized cuts, thickness gradients, or offset curvatures can reshape the energy landscape and guide the snapping direction. For example, Li et al.~\cite{Li_2024} demonstrated a semi-open elastomeric dome with a T-cut that snapped consistently in one direction under pneumatic pressure. Bell et al.~\cite{bell_cattani_gorissen_bertoldi_weaver_wood_2020} introduced a thickness gradient in a bistable snapping shell to create an asymmetric snapping actuator, enabling mechanically programmed complex gaits under simple periodic fluidic input.

Material and actuation asymmetries further enhance the physical control. Local stiffness variation and sequenced inputs guide deformation paths and transition timing. Luo et al.~\cite{Luo_2024} created a multistable soft actuator where internal asymmetries drove mixed-mode snapping between stable states. These behaviors enable fast, reconfigurable motion governed by embedded physical design rather than external control. 

In this work, as shown in \mbox{Fig.\ \ref{fig:fig0}}, we proposed an eccentric snapping actuator capable of generating nonreciprocal asymmetric motion. To evaluate its performance, we conducted both experimental motion tracking and finite element simulations to validate the deformation behavior and the snapping sequence. Furthermore, a sequencing actuation system was developed to integrate the actuator into a bio-inspired walking robot, enabling coordinated movements similar to those observed in natural organisms. Finally, the locomotion performance of the robot was characterized under different actuation frequencies ranging from 1 Hz to 7.5 Hz, demonstrating the potential of the proposed design for energy-efficient and physically intelligent soft robotic systems.

\section{Design and Methods}

\subsection{Actuator Design}
In studies of snapping shells, most attention has been devoted to the nonlinear behavior during the snap-through transition. However, the asymmetric motion that accompanies this process is also significant, as similar asymmetric deformation is commonly found in biological systems~\cite{nixonHumanIdentificationBased2010}. During fabrication of snapping shell (e.g. like the design reported in \cite{van_raemdonck_milana_de_volder_reynaerts_gorissen_2023}), it can be observed that inevitable geometric imperfections introduced by the manufacturing process play a critical role in determining the snapping deformation path. Although the shell is designed with uniform thickness, achieving complete uniformity in practice is very challenging. Consequently, one region of the shell tends to be slightly thinner and therefore softer than the other regions. Experimental observation shows that the snap-through process consistently initiates from this softer region and then propagates toward the stiffer region, resulting in an asymmetric deformation sequence.

Inspired by this observations and to intentionally amplify and control this asymmetry, an eccentric snapping shell design was developed and introduced in this work. The concept is to weaken one side of the shell by introducing an offset in the dome geometry. As shown in \mbox{Fig.\ \ref{fig:fig2}}, the geometric center of the dome was shifted by \SI{57}{\percent}, which is 2 mm relative to the base circle, which made the left side weaker due to its smaller local curvature angle. The shell thickness was also reduced to decrease the strain energy barrier required for snap-through, thereby enabling actuation under lower pneumatic pressure (e.g. \( 40~\text{mbar} \)).

\begin{figure}[H]
    \centering
    \includegraphics[width=0.4\textwidth]{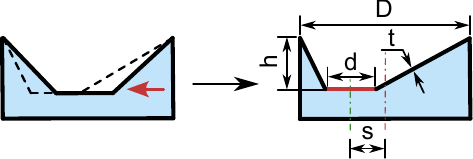}
    \caption{Eccentric structure design of asymmetric snapping actuator.}
    \label{fig:fig2}
\end{figure}

For pneumatic operation, the shell was integrated with a support chamber of 3.5 mm depth. When compressed air is introduced, the internal pressure within the chamber increases and drives the snap-through motion. As the pressure rises, the weaker left side deforms first, causing the central region of the shell to tilt toward the right. With further pressurization, the right side subsequently deforms, restoring the central region to an approximately horizontal configuration. During depressurization, the sequence reverses: the left side recovers first, tilting the central region to the left, followed by recovery of the right side, which returns the shell to its initial position. This sequential deformation behavior results in a complete cycle of asymmetric motion.

\subsection{Robot Design}
\begin{figure*}[!b]
    \setcounter{figure}{4}
    \centering
    \includegraphics[width=0.80\textwidth]{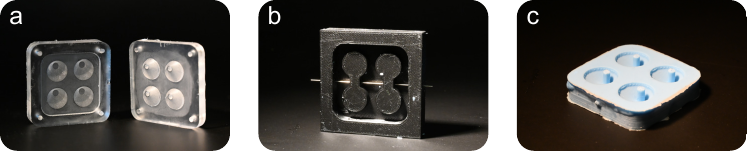}
    \caption{Fabrication process: a) mold for actuator fabricated by milling machine, b) mold for actuation system fabricated by FDM 3D printer, c) the final assembled robot.}
    \label{fig:fig5}
\end{figure*}
With the asymmetric actuator, it becomes possible to replicate the locomotion patterns observed in natural organisms. In biological systems, stable walking is often achieved through the coordinated out-of-phase motion of multiple limbs~\cite{dickinsonHowAnimalsMove2000}. To emulate this mechanism,  we design an integrated four-legged soft robot, where a sequencing actuation pattern is physically controlled by integrating four asymmetric snapping actuators to realize phase-shifted actuation for stable quadrupedal locomotion, using only a single pressure input.

\begin{figure}[H]
    \setcounter{figure}{2}
    \centering
    \includegraphics[width=0.4\textwidth]{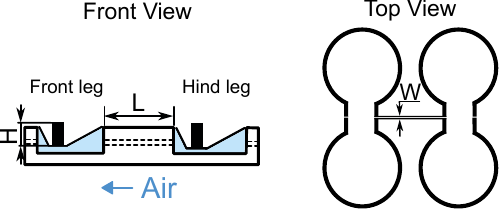}
    \caption{Sequential pneumatic network system design.}
    \label{fig:fig3}
\end{figure}

The parameters of the asymmetric shell and robot are summarized in Table 1.

\begin{table}[h!]
\centering
\caption{Geometric parameters of actuator}
\label{tab:design_parameters}
\begin{tabular}{lcc}
\hline
\textbf{Parameter} & \textbf{Symbol} & \textbf{Value(mm)} \\
\hline
Shell height & $h$ & 3 \\
Tip diameter & $d$ & 3 \\
Shell diameter & $D$ & 10 \\
Shell thickness & $t$ & 0.2 \\
Eccentric offset & $s$ & 2 \\
Channel width & $W$ & 0.4 \\
Channel length & $L$ & 9 \\
Tip length & $H$ & 5 \\
\hline
\end{tabular}
\end{table}

As illustrated in \mbox{Fig.\ \ref{fig:fig3}}, the four actuators were divided into two groups. Within each group, the actuators were directly connected, while the two groups were linked through a narrow pneumatic channel. When air is injected into the chamber, the flow resistance in the narrow channel introduces a pressure delay between the two groups, resulting in sequential actuation, an effect that was already explored in ~\cite{gorissenHardwareSequencingInflatable2019,milanaMorphologicalControlCiliaInspired2022}. The group located near the air inlet functions analogously to the hind legs, while the distal group behaves as the forelegs. Additionally, an air outlet was installed near the foreleg group to facilitate pressure release and enhance cyclic operation.

\begin{figure}[H]
    \centering
    \includegraphics[width=0.35\textwidth]{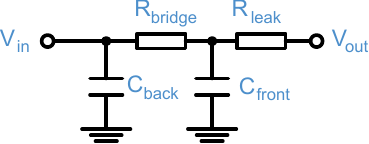}
    \caption{RC circuit analogy for the sequential pneumatic network.}
    \label{fig:fig4}
\end{figure}

This pneumatic network embodying the physical control sequencing can be conceptually represented by an RC circuit analogy, as shown in \mbox{Fig.\ \ref{fig:fig4}}. In this model, the main chambers correspond to capacitors (C) that store energy, while the narrow channels and outlets act as fluidic resistive elements (R) governing flow delay. When the bridge resistance (R\textsubscript{bridge}) is large, corresponding to a very narrow channel, the rear chamber (C\textsubscript{back}) inflates first, followed by the front chamber (C\textsubscript{front}), producing a time-delayed sequential actuation. By tuning the effective resistance and capacitance parameters, the timing and phase of actuation can be adjusted to achieve different gait patterns.

To further amplify the locomotion output, as shown in Figure 3, a 5 mm pillar was attached to the central region of each snapping shell, serving as a contact leg. This design enhances the vertical displacement and improves ground contact efficiency during walking.

\subsection{Fabrication}
To fabricate the snapping shell, precise control of shell thickness was required. Therefore, a Carvera Air Desktop CNC milling machine was employed to machine the mold. Acrylic plates were used as the mold material. Flat-end milling tools (3.175 × 25 mm and 1 × 4 mm) were applied for rough machining, followed by a 30° × 0.2 mm engraving tool for fine finishing. The resulting mold is shown in \mbox{Fig.\ \ref{fig:fig5}}a. Subsequently, Smooth-Sil 950 silicone was injected into the mold to form the shell structure.

For the pneumatic network, the mold was fabricated using a PRUSA MK4 3D printer with ABS filament. A 0.4 mm stainless steel rod was inserted into the mold to create the internal connecting microchannel. The corresponding mold is shown in \mbox{Fig.\ \ref{fig:fig5}}b. After molding, Smooth-Sil 950 silicone was poured into the cavity and cured at 70 °C for 1 hour.

After all parts were fabricated, they were bonded together using the same silicone material and subjected to a secondary curing process at 70 °C for 1 hour to ensure airtight sealing and mechanical integrity. The final assembled actuator is shown in \mbox{Fig.\ \ref{fig:fig5}}c. For pneumatic connection, a 1 mm-diameter stainless steel needle was inserted at the inlet to serve as the air connector.

\subsection{Finite element simulations}
A numerical simulation of the inflation--deflation process was conducted in \textit{Abaqus/CAE 2025} using a fluid--structure interaction (FSI) framework with a \texttt{*FLUID CAVITY} definition. A dynamic implicit analysis was performed to capture the deformation evolution under controlled volumetric loading and to evaluate the corresponding pressure--volume response.

The membrane was modeled as a Neo--Hookean hyperelastic material~\cite{xavier_fleming_yong_2020,gariyaStressBendingAnalysis2022} with shear constant \( C_{1} = 0.34~\text{MPa} \). Rayleigh damping was introduced with coefficients \( \alpha = 0.0001 \) and \( \beta = 0.0001 \) to suppress numerical oscillations. Both the inflation and deflation stages lasted \( 1~\text{s} \), using a Dynamic, Implicit step with nonlinear geometric effects enabled. The time increment was limited between \( \Delta t_{\min} = 1\times10^{-5}~\text{s} \) and \( \Delta t_{\max} = 1\times10^{-4}~\text{s} \).

A prescribed fluid flux was implemented through the \texttt{*FLUID FLUX} function to apply a ramp volumetric flow rate with a peak value of \( 0.4~\text{mL}\,\text{s}^{-1} \). The base plane was fully constrained to replicate the experimental boundary conditions, where the region was bonded to a \( 5~\text{mm} \) support bar. The structure was discretized using S4R reduced-integration shell elements with a global mesh size of approximately \( 0.3~\text{mm} \).
\begin{figure*}[t]
    \setcounter{figure}{5}
    \centering
    \includegraphics[width=0.8\textwidth]{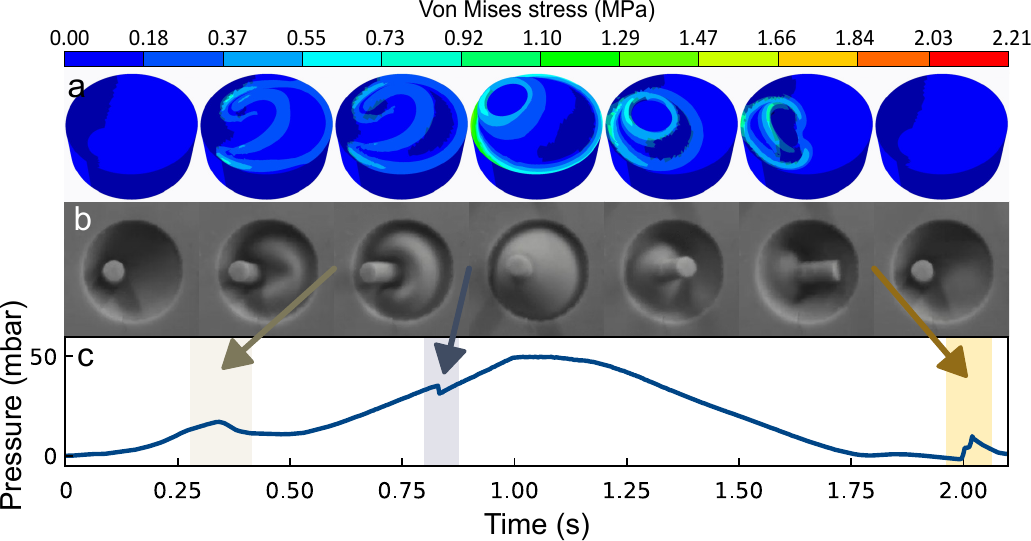}
    \caption{Snapping process of the asymmetric actuator: a) simulation result of the snapping process, b) experimental result of the snapping process, c) pressure curve of the snapping process. Pictures are adapted from Supporting Movie.}
    \label{fig:fig8}
\end{figure*}
\subsection{Experimental Setup}

\subsubsection{Pressure curve} 
To measure the pressure response of the snapping shell, a syringe pump (Harvard Apparatus PHD Ultra) was used to inject air into the actuator at a constant rate of \( 0.4~\text{mL}\,\text{s}^{-1} \), with a target volume of \( 0.4~\text{mL} \). The internal pressure of the actuator was monitored using a pressure sensor (Honeywell ABPDRRV060MGAA5). Data acquisition and synchronization were performed using a DAQ system (National Instruments USB{-}6212).

    

\subsubsection{Trajectory analysis}
To capture the nonreciprocal deformation behavior of the shell during actuation and track the trajectory of the pillar, a high--speed camera (Phantom Miro C321) was employed, recording at 600 fps.

\subsubsection{Actuation}
For generating oscillatory pneumatic excitation, a proportional pressure regulator (Festo VPPI{-}5L{-}3{-}G18{-}1V1H{-}V1{-}S1D) was utilized. The actuation signal was provided by a waveform generator outputting a ramp waveform with an amplitude of \( 2{-}8~\text{V} \) and frequencies ranging from \( 1~\text{Hz} \) to \( 5~\text{Hz} \).

\section{Results}
\begin{figure*}[t]
    \setcounter{figure}{7}
    \centering
    \includegraphics[width=0.89\textwidth]{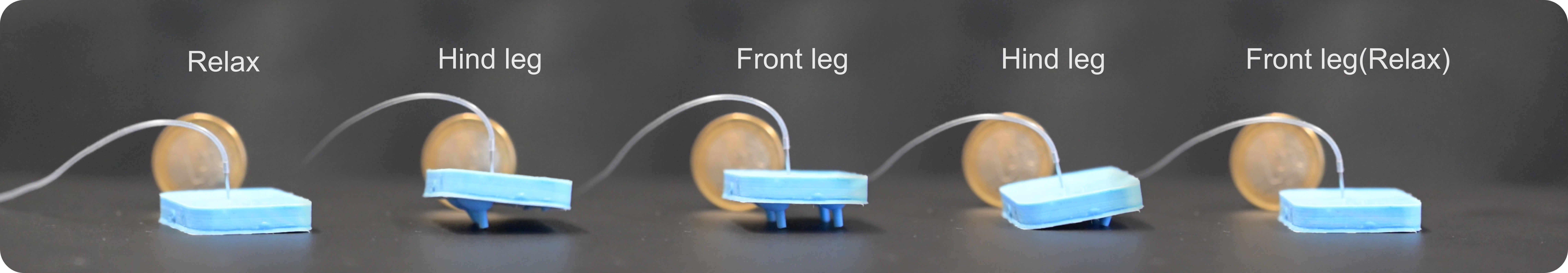}
    \caption{Breakdown walking process of the robot under 1Hz.}
    \label{fig:fig1}
\end{figure*}
\subsection{Deformation and Pressure Curve}

As shown in \mbox{Fig.\,\ref{fig:fig8}}a and \mbox{Fig.\,\ref{fig:fig8}}b, the snapping process of the actuator can be divided into seven phases, with the snap-through state considered as the symmetric reference. These phases describe the full deformation cycle from inflation to deflation.

During inflation (0--1 s), the snapping shell exhibits two characteristic states. The first corresponds to the onset of geometric instability, highlighted in the brown region in \mbox{Fig.\,\ref{fig:fig8}}c. Due to geometric asymmetry, the shell loses stability and undergoes a rapid shape transition, producing a small bump in the pressure curve. The second is the snap-through event, which occurs when the internal pressure reaches the critical threshold of approximately 36 mbar. At this point, a sudden deformation of the shell leads to a sharp pressure drop, as shown in the purple region in \mbox{Fig.\,\ref{fig:fig8}}c. Throughout the inflation stage, the strain on the right side remains higher than that on the left, consistent with the eccentric design. The weaker right side responds more sensitively to pressure changes, causing the central region of the shell to tilt leftward before fully snapping through.

After inflation, the syringe pump withdraws to reduce the internal pressure. However, a short switching delay of approximately 0.1~s (\(\sim0.1~\text{s}\)) occurs between injection and withdrawal due to the mechanical design of the pump. As a result, the total actuation period extends to about 2.1~s instead of the nominal 2~s.

During deflation (\(1\text{--}2~\text{s}\)), the shell undergoes a snap-back transition, highlighted in the yellow region in \mbox{Fig.\,\ref{fig:fig8}}c. The snap-back threshold is approximately \(0~\text{mbar}\). In this phase, the center tilts rightward, and the strain on the left side becomes higher, as the weaker region responds first to pressure release. This premature recovery compresses the left side, explaining the observed strain asymmetry. Over a full cycle, the actuator completes one asymmetric snapping motion.

\begin{figure}[H]
    \setcounter{figure}{6}
    \centering
    \includegraphics[width=0.4\textwidth]{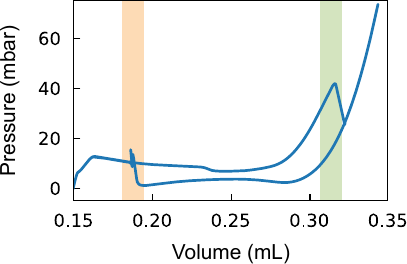}
    \caption{Finite element simulation of the pressure-volume curve.}
    \label{fig:fig10}
\end{figure}

As shown in \mbox{Fig.\,\ref{fig:fig10}}, the pressure--volume curve reveals clear energy release during the process, which explains why the snap-back threshold is lower than the snap-through threshold. Numerical simulations further confirm that the snap-through threshold, indicated in the green region, occurs at approximately 41 mbar, while the snap-back threshold, indicated in the orange region, is near 0 mbar, showing good agreement with the experimental results.

To quantify the hysteresis ratio, we integrate the pressure--volume (PV) curve:
\[
H=\frac{W_{\mathrm{in}}-W_{\mathrm{out}}}{W_{\mathrm{in}}}
=\frac{\int_{\mathrm{loading}} P\,dV-\int_{\mathrm{unloading}} P\,dV}{\int_{\mathrm{loading}} P\,dV}\]
\[
=\frac{1.171}{3.201}=36.6\%.
\]

Thus, 36.6\% of the input energy is dissipated over one cycle.

\subsection{Trajectory}
As shown in \mbox{Fig.\,\ref{fig:fig11}}a, the simulation predicts an actuator trajectory with an $\sim$8\,mm displacement along the $x$-axis and an $\sim$5\,mm displacement along the $y$-axis. Compared with the experimental result in \mbox{Fig.\,\ref{fig:fig11}}b, the simulated and measured trajectories agree well overall. The swept area is $\approx 28.647\,\mathrm{mm}^2$ in simulation and $28.144\,\mathrm{mm}^2$ in the experiment.

The remaining discrepancies mainly stem from the supporting pillar model. In the experiment, the pillar is soft silicone, whereas in the simulation it is represented as a reference point without stiffness or thickness. As a result, small tip vibrations appear experimentally but not in simulation. In addition, the simulated trajectory is slightly more extreme at the lower-right region because the pillar has no thickness in the model and thus does not experience edge blocking. In contrast, the real pillar (2\,mm diameter) is constrained at extreme positions.

These results confirm the controllable asymmetric snapping behavior and validate the effectiveness of the proposed eccentric design.

\begin{figure}[H]
    \setcounter{figure}{8}
    \centering
    \includegraphics[width=0.49\textwidth]{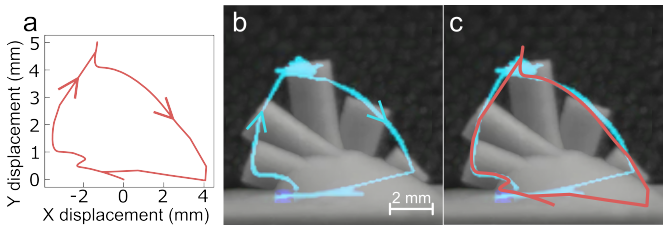}
    \caption{Results of trajectory of the asymmetric actuator: a) simulation result, b) experimental result under high speed camera., c) strong agreement between experimental tracking and simulation}
    \label{fig:fig11}
\end{figure}

\subsection{Walking Performance}
\subsubsection{Movement Breakdown}
As shown in \mbox{Fig.\,\ref{fig:fig1}}, during a complete actuation cycle, the hind legs are activated first during inflation, followed by the front legs. During deflation, the sequence reverses. As we can see in the figure, the hind legs retract first, then the front legs, returning the robot to its relaxed state. This coordinated sequence produces a distinct wavelike locomotion pattern, clearly demonstrating the effectiveness of the proposed pneumatic sequencing network. The robot has a compact size of approximately 25~mm on each side, comparable to the diameter of a one-euro coin. At an actuation frequency of 1~Hz, the robot achieves a forward stride of approximately 5~mm per cycle. Due to its simple structure and fully integrated actuation system, the design can potentially be further miniaturized for future applications.

\subsubsection{Locomotion at different frequencies}
\begin{figure}[htbp]
    \setcounter{figure}{9}
    \centering
    \includegraphics[width=0.4\textwidth]{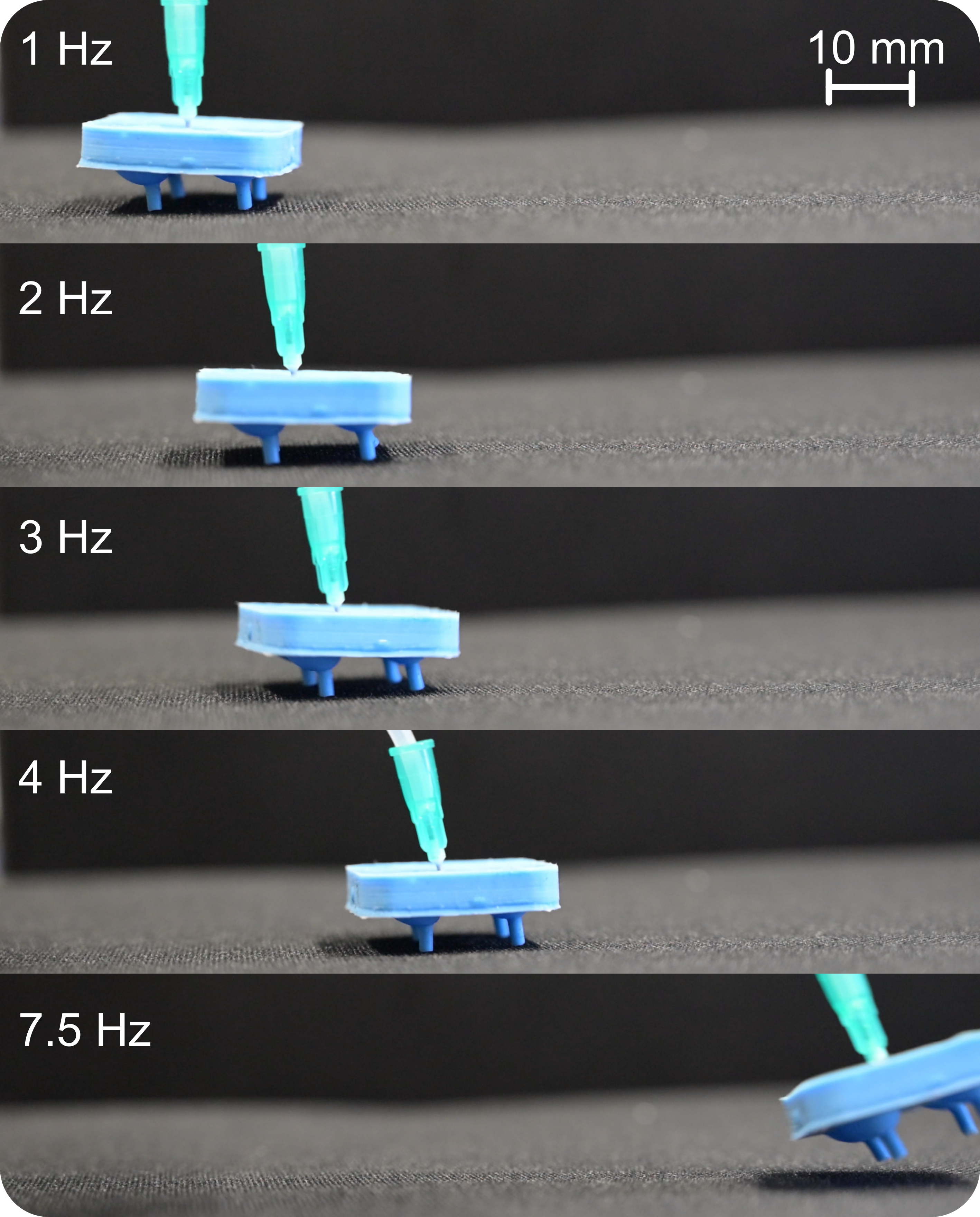}
    \caption{Forward movement under different frequencies. Pictures adapted from Supporting Movie.}
    \label{fig:fig12}
\end{figure}
As shown in \mbox{Fig.\,\ref{fig:fig12}}, the robot was actuated under an oscillatory pressure of approximately \(\pm 600~\text{mbar}\) at different frequencies. The results indicate a clear frequency-dependent locomotion performance: the average speed increased from 5.13~mm\,s\(^{-1}\) at 1~Hz to 10.94~mm\,s\(^{-1}\) at 2~Hz, 14.21~mm\,s\(^{-1}\) at 3~Hz, 16.83~mm\,s\(^{-1}\) at 4~Hz, and 72.78~mm\,s\(^{-1}\) at 7.5~Hz. These correspond to normalized speeds of approximately \SI{0.21}{\bodylength\per\second}, \SI{0.44}{\bodylength\per\second}, \SI{0.57}{\bodylength\per\second}, \SI{0.78}{\bodylength\per\second}, and \SI{2.91}{\bodylength\per\second}, respectively.

The sharp rise in velocity at 7.5~Hz indicates a transition from a walking gait to a jumping-like motion. As shown in phases 3 and 7 of \mbox{Fig.\,\ref{fig:fig13}}, the snapping dome remains asymmetric at high frequency, but the motion is dominated by the hind leg. The higher actuation rate strengthens the hind-leg snap, causing the robot to lift off and remain airborne briefly. Meanwhile, increased damping between actuators at high frequency suppresses actuation of the front leg, which instead serves as a passive support, analogous to a kangaroo's tail.

These results demonstrate that the locomotion mode and speed of the robot can be effectively tuned by controlling the frequency of pneumatic actuation.

\begin{figure}[htbp]
    \setcounter{figure}{10}
    \centering
    \includegraphics[width=0.45\textwidth]{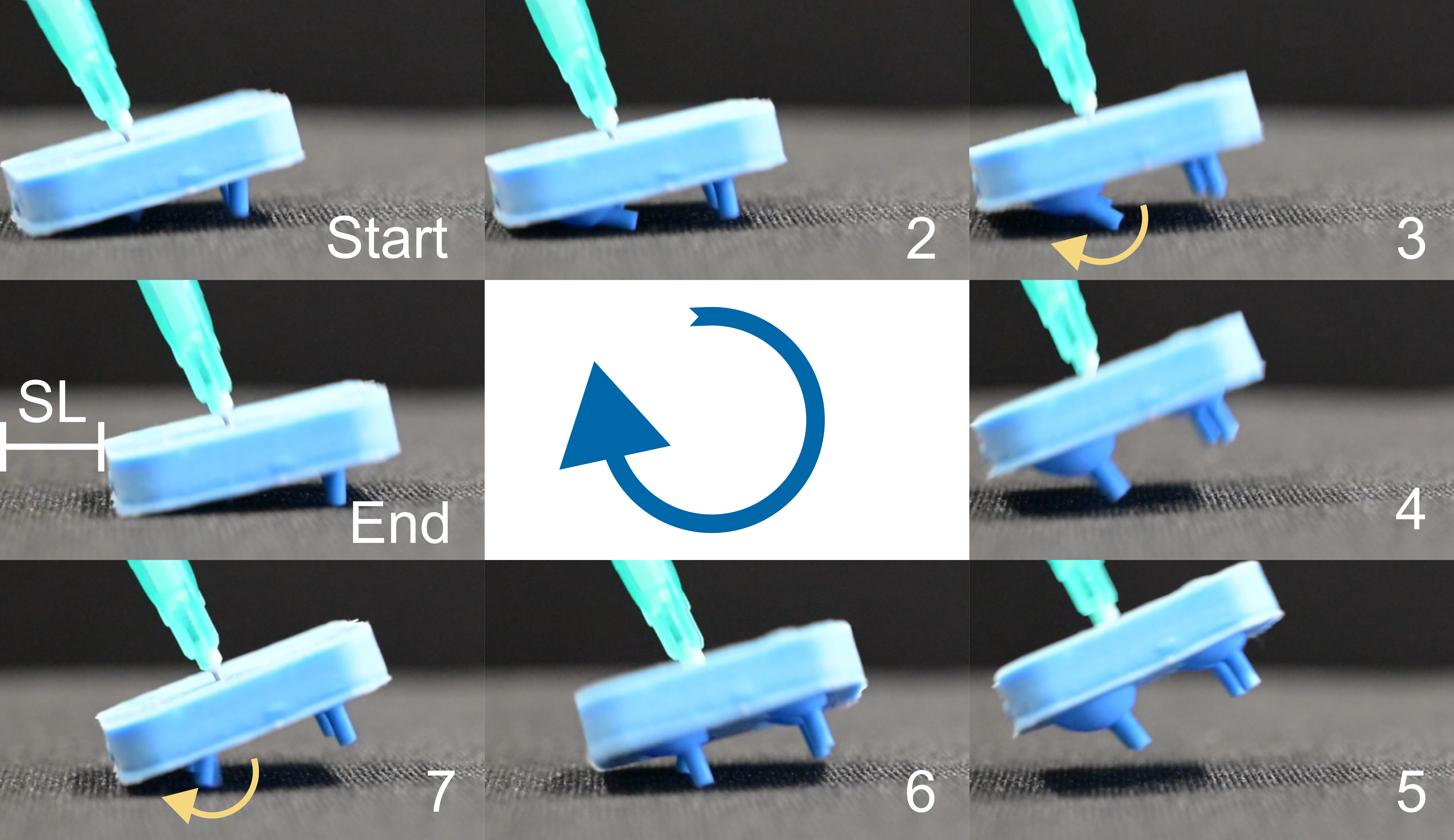}
    \caption{Gait-phase diagram of the jumping-like motion under 7.5Hz}
    \label{fig:fig13}
\end{figure}


\section{Conclusions}

In this work, we presented an eccentric snapping actuator capable of generating controllable asymmetric motion through geometry-induced instability. By introducing a deliberate offset in the dome structure, the actuator exhibits a directional snap-through and snap-back sequence, which was validated through both finite element simulations and experimental measurements. The pressure–volume analysis revealed strong agreement between simulation and experiment, confirming the nonlinear mechanical behavior and associated energy release during each actuation cycle.

Building upon this principle, a sequencing pneumatic network was developed to coordinate multiple asymmetric actuators into a compact four-legged soft robot. The integrated design enables physical control as a sequential actuation using a single pressure input. The robot demonstrated frequency-dependent locomotion performance, reaching a maximum speed of \SI{2.91}{\bodylength\per\second} at 7.5~Hz, and exhibiting a transition from walking to jumping-like motion at higher frequencies.

Overall, these results highlight the potential of geometry-guided snapping structures for realizing physically controlled actuation and locomotion. The proposed design simplifies pneumatic control and enables dynamic adaptive motion, thus offering a promising route toward compact, energy-efficient, and self-regulating soft robotic systems.

In future work, we plan to integrate pressure sources \cite{wehner_truby_fitzgerald_mosadegh_whitesides_lewis_wood_2016} and embodied pressure-regulation components \cite{van_laake_de_vries_malek_kani_overvelde_2022,rothemund_ainla_belding_preston_kurihara_suo_whitesides_2018} directly into the robot body to achieve a completely untethered system. Furthermore, extended numerical studies will be conducted to optimize the geometric parameters of the snapping shell for enhanced stability and controllability of the asymmetric motion. Additional simulations of the pneumatic sequencing network, e.g., using the FONS framework reported by Baeyens et al.~\cite{baeyens_van}, will be conducted to identify parameter combinations that broaden the frequency response and further improve locomotion performance.


\section*{Acknowledgment}
This research was Funded by the Deutsche Forschungsgemeinschaft (DFG, German Research Foundation) under Germany’s Excellence Strategy – EXC-2193/1 – 390951807. Additional funding was provided by the Volkswagen Foundation - 9D761.

\bibliographystyle{ieeetr}
\bibliography{references}
\end{document}